\documentclass[11pt]{article}

\usepackage[margin=1in]{geometry}
\usepackage[T1]{fontenc}
\usepackage[utf8]{inputenc}

\usepackage{newpxtext,newpxmath}

\usepackage{amsmath}
\usepackage{graphicx}
\usepackage{booktabs}
\usepackage{placeins}
\usepackage{caption}
\usepackage[numbers,sort&compress]{natbib}

\usepackage[dvipsnames]{xcolor}
\definecolor{TitleColor}{RGB}{28,37,54}
\definecolor{SectionColor}{RGB}{55,78,107}
\definecolor{LinkColor}{RGB}{74,102,136}
\definecolor{RuleColor}{RGB}{185,188,192}

\usepackage[
    colorlinks=true,
    linkcolor=LinkColor,
    citecolor=LinkColor,
    urlcolor=LinkColor
]{hyperref}

\usepackage{authblk}

\usepackage{setspace}
\usepackage{titlesec}
\usepackage{needspace}

\titleformat{\section}
  {\large\bfseries\color{SectionColor}}
  {\thesection.}{0.5em}{}
\titleformat{\subsection}
  {\normalsize\bfseries\color{SectionColor}}
  {\thesubsection}{0.5em}{}
\titleformat{\subsubsection}
  {\normalsize\bfseries\color{SectionColor}}
  {\thesubsubsection}{0.5em}{}
\titleformat{\paragraph}[runin]
  {\normalsize\bfseries\color{SectionColor}}
  {}{0em}{}
\titlespacing*{\paragraph}{0pt}{0.9em}{0.5em}

\usepackage{fancyhdr}
\fancypagestyle{firstpage}{
    \fancyhf{}
    \renewcommand{\headrulewidth}{0pt}
    \cfoot{\thepage}
}

\renewcommand{\headrulewidth}{0.4pt}
\renewcommand{\headrule}{\hbox to\headwidth{\color{RuleColor}\leaders\hrule height \headrulewidth\hfill}}

\title{\textbf{\color{TitleColor} Toward Auditable AI Scientists: \\ A Hypothesis Evolution Protocol for LLM Agents}}
\author[1, *]{Izumi Takahara}
\author[1, $\dagger$]{Teruyasu Mizoguchi}
\affil[1]{Institute of Industrial Science, The University of Tokyo, Tokyo, Japan 153-8505}
\affil[*]{kougen@iis.u-tokyo.ac.jp, \quad \textsuperscript{$\dagger$}teru@iis.u-tokyo.ac.jp}
\date{}

\begin{document}

\maketitle
\thispagestyle{firstpage}

\begin{quote}
\small\noindent
Large language model (LLM) agents are increasingly expected to play a
central role in AI-driven scientific discovery.
Equipped with broad knowledge, flexible reasoning, and tool use, they
have the potential to autonomously explore and solve scientific
problems by repeatedly proposing hypotheses, testing them, and
revising their beliefs in the light of the evidence.
In current agents, however, these hypotheses, tests, and belief updates
are buried in unstructured logs, and no mechanism lets the agent or the
human researcher audit that process.
Here we propose the Hypothesis Evolution Protocol (HEP), an agent
harness that provides hypothesis generation, evaluation, and evolution
as explicit, auditable operations.
On materials-science research tasks, a HEP-equipped agent operates the
hypothesis--test--evidence--belief cycle that planning-style agents lack,
generalizes across research questions, and exploits the protocol more
fully as the base LLM becomes more capable.
These results mark a step toward auditable AI scientists, whose
scientific reasoning can be inspected, verified, and built upon.
\end{quote}


\section{Introduction}

The rapid progress of large language models (LLMs), and in particular of
LLM-based agents, has attracted intense interest as a potential driver of
scientific discovery across a wide range of
domains~\cite{Lu2026AIScientist,Gottweis2026CoScientist,Mitchener2025Kosmos,Zhang2026AgenticMaterials}.
Modern LLMs combine broad scientific knowledge with flexible reasoning
capabilities, and when equipped with
tools~\cite{Yao2023ReAct,Anthropic2024MCP,Hou2025MCP},
skills~\cite{Wang2023Voyager,Jiang2026AgenticSkills},
memory~\cite{Chhikara2025Mem0,Gao2026SelfEvolvingAgents}, and reasoning strategies such
as planning and
reflection~\cite{Yao2023ReAct,Sun2023AdaPlanner,Shinn2023Reflexion},
the resulting
agents have the potential to drive scientific research flexibly and
autonomously.

To date, LLM agents have been developed toward the autonomous execution of
research in several fields.
Prominent examples include the end-to-end automation of artificial
intelligence research and engineering~\cite{Lu2026AIScientist,Jin2026Arbor}
and the automation of biomedical
discovery~\cite{Gottweis2026CoScientist,Ghareeb2025Robin,Mitchener2025Kosmos,Ke2025BioDisco},
as well as chemistry agents that couple LLMs with domain tools and robotic
laboratories~\cite{Bran2024ChemCrow,Boiko2023Coscientist,Darvish2025ORGANA}.
In materials science, LLM agents have been actively applied both to
computational materials
design~\cite{Jia2024LLMatDesign,Takahara2025MatAgent,Nduma2025Crystalyse,Kim2026Materealize,Zou2025ElAgente,Pham2026ChemGraph,Deng2026AtomisticSkills}
and to the automation of materials synthesis and
characterization~\cite{Fei2026AgenticLLM,YanguasGil2026ALDOptimization,shi2026knowledge,mandal2025evaluating,vriza2026operating},
and they are emerging as a practical route to accelerating the progress
of materials research~\cite{Zhang2026AgenticMaterials}.

The term ``autonomous research'' by LLM agents, however, encompasses
tasks with substantially different levels of
agency~\cite{RiosGarcia2026AIScientists}.
Much of the work above automates predefined workflows, in which the agent
reliably executes a prescribed sequence of computations, syntheses, or
measurements~\cite{Bran2024ChemCrow,Boiko2023Coscientist,Darvish2025ORGANA,Deng2026AtomisticSkills},
or performs optimization toward an externally specified
goal~\cite{YanguasGil2026ALDOptimization,Fei2026AgenticLLM}.
A task demanding substantially higher agency is to answer the question of
\emph{why}: to arrive at explanatory understanding that is earned through
repeated cycles of generating mechanistic hypotheses, devising the
empirical tests that can discriminate among them, and revising beliefs
against the resulting evidence.
Steps in this direction include agents that generate and rank scientific
hypotheses~\cite{Ghafarollahi2025SciAgents,Yang2024MOOSE,Yang2025MOOSEChem,Yang2025MOOSEChem2,Ke2025BioDisco,Gottweis2026CoScientist},
frameworks that place falsification at the center of the research
process~\cite{Liu2024AIGS,Huang2025POPPER}, belief-guided open-ended
exploration~\cite{Agarwal2025AutoDiscovery}, and closed-loop systems that
iteratively refine hypotheses against data, experiments, or
simulations~\cite{Mitchener2025Kosmos,Ghareeb2025Robin,Wang2026ScienceClaw,Jin2026Arbor,Wiemann2026DiscoverPhysics}.

Nevertheless, in existing agents, neither the agent itself nor the
human researcher can readily trace which hypotheses the agent
entertained, which tests it performed, and how the resulting evidence
moved its beliefs.
Hypotheses and beliefs typically remain implicit in free-text
reasoning or in undisclosed internal states.
Where the research record is externalized, what is tracked is
argument-based rankings~\cite{Gottweis2026CoScientist}, the provenance
of artifacts~\cite{Wang2026ScienceClaw}, or task-performance
scores~\cite{Jin2026Arbor}, rather than an explicit degree of belief
that empirical evidence can raise or lower.
Indeed, a recent large-scale analysis of agent execution traces found that
evidence was ignored in about two-thirds of traces and that beliefs were
revised after refutation in only about a quarter of
cases~\cite{RiosGarcia2026AIScientists}.
Without an explicit, auditable record of the
hypothesis--test--evidence--belief cycle, and without a structure that
disciplines how the cycle is operated, the scientific soundness of an
agent's conclusions cannot be
verified~\cite{Wang2026WorkflowClosure,Bisht2026AgenticNotBuilt}, and
the agent itself cannot systematically build on, revise, or retire its
own hypotheses over a long research horizon.

In this work, we propose the \emph{Hypothesis Evolution Protocol} (HEP),
an agent harness that turns the scientific procedure of proposing,
testing, and revising hypotheses in the light of evidence into an
explicit protocol, every step of which is traceable by both the LLM
agent and the human researcher.
HEP externalizes each hypothesis as a persistent object in an auditable
registry, where it carries a belief probability that can be moved
only by attached evidence, evolves through refinement and merging of parent
hypotheses, and progresses through a lifecycle from proposal to
resolution as supported, refuted, or dormant.
The protocol thus serves as a footing on which the agent operates the
hypothesis--test--evidence--belief cycle in a disciplined way.
Using materials-science research tasks as a testbed, we show that an agent
equipped with HEP actually operates the hypothesis--test--evidence--belief
cycle that planning-style agents lack, that the same protocol generalizes
across research questions, and that the depth
to which the protocol is exploited scales with base-LLM capability.

\begin{figure}[p]
    \centering
    \includegraphics[width=0.95\textwidth]{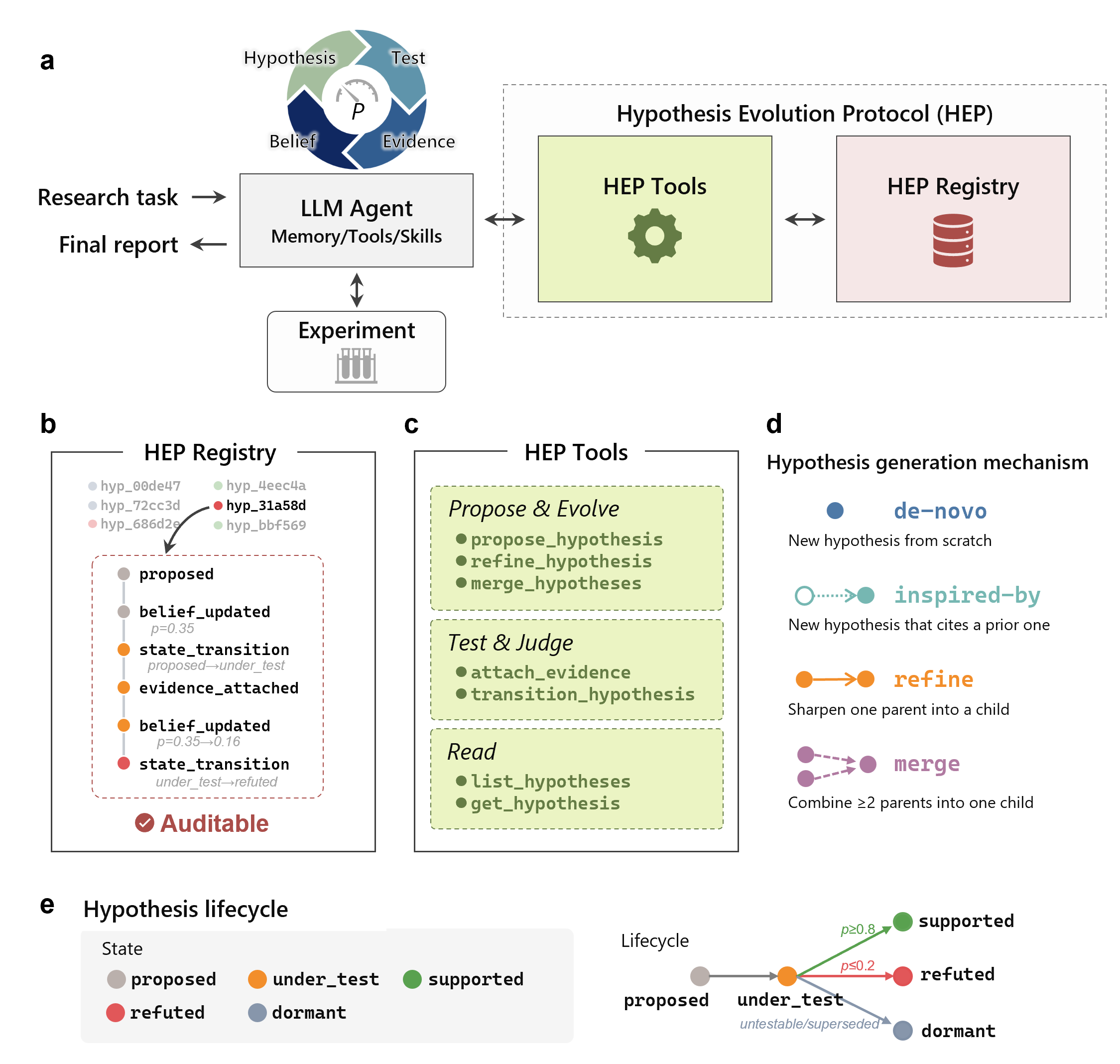}
    \caption{%
        \textbf{Overview of the Hypothesis Evolution Protocol (HEP).}
        \textbf{(a)} Given a research task, an LLM agent runs experiments
        and any other tools as usual, while managing its hypotheses,
        evidence, and beliefs through HEP.
        The protocol guides the agent
        through iterations of the Hypothesis--Test--Evidence--Belief cycle,
        and the task concludes with a final report.
        HEP consists of an auditable \emph{HEP Registry} and
        \emph{HEP Tools}.
        \textbf{(b)} The HEP Registry is the store that manages the state of
        every hypothesis.
        Each is kept under a unique hypothesis id as a persistent object
        with an append-only event log.
        The expanded example shows the event log of hypothesis
        \texttt{hyp\_31a58d}, which is proposed with prior belief
        $P(H)=0.35$ and refuted after an attached piece of evidence lowers
        its belief to $0.16$.
        Every event remains in the append-only log, so the full record is
        auditable.
        \textbf{(c)} HEP Tools expose the HEP Registry to the agent in
        three groups: \emph{Propose \& Evolve}, \emph{Test \& Judge}, and
        \emph{Read}.
        \textbf{(d)} Hypotheses are generated and evolved by four mechanisms:
        \emph{de-novo}, \emph{inspired-by}, \emph{refine}, and \emph{merge}.
        \textbf{(e)} A hypothesis managed through HEP takes one of five
        lifecycle states: \texttt{proposed}, \texttt{under\_test},
        \texttt{supported}, \texttt{refuted}, or \texttt{dormant}.
        State transitions are performed through the
        \texttt{transition\_hypothesis} tool, and every hypothesis is
        expected to be settled as \texttt{supported}, \texttt{refuted},
        or \texttt{dormant}.
    }
    \label{fig:hep_overview}
\end{figure}

\section{Results}

\subsection{Hypothesis Evolution Protocol}
Figure~\ref{fig:hep_overview} presents an overview of the proposed
HEP.
An LLM agent tackling a research task interacts with its experimental
environment and any other tools as usual.
On top of this ordinary
workflow, HEP provides the layer through which the agent manages its
hypotheses, evidence, and beliefs (Fig.~\ref{fig:hep_overview}a).
The protocol consists of two components: an auditable
\emph{HEP Registry} and a set of \emph{HEP Tools}.
The HEP Registry records each hypothesis as a persistent object with a
unique hypothesis id, carrying a natural-language statement, a belief
$P(H)$, defined as the probability the agent currently assigns to the
hypothesis being true, a lifecycle state, and its provenance.
Every belief update, evidence attachment, and state transition is
appended to the event log of the corresponding hypothesis, so the full
history of each hypothesis remains auditable to both the agent and the
human researcher (Fig.~\ref{fig:hep_overview}b).

HEP Tools mediate the interaction between the agent and the HEP
Registry in three groups, \emph{Propose \& Evolve}, \emph{Test \&
Judge}, and \emph{Read}, the last of which lets the agent look up the
tracked hypotheses and their current states and beliefs
(Fig.~\ref{fig:hep_overview}c).
Hypotheses enter the registry through four generation mechanisms,
\emph{de-novo}, \emph{inspired-by}, \emph{refine}, and \emph{merge}
(Fig.~\ref{fig:hep_overview}d).
As the agent invokes these tools over
the course of a run, its hypotheses evolve, and their population comes
to form an explicit lineage rather than a flat list.

Each hypothesis takes one of five lifecycle states, from
\texttt{proposed} and \texttt{under\_test} to a resolution as
\texttt{supported}, \texttt{refuted}, or \texttt{dormant}
(Fig.~\ref{fig:hep_overview}e).
Belief is disciplined by two rules.
First, $P(H)$ can be moved only by evidence that passes a validation
gate.
Attachments judged insufficient are recorded but leave belief
unchanged.
Second, verdicts are threshold transitions enforced by the registry:
a hypothesis can be transitioned to \texttt{supported} only when
$P(H)\ge 0.8$ and to \texttt{refuted} only when $P(H)\le 0.2$, while
hypotheses that cannot be tested further are shelved as
\texttt{dormant} with an explicit reason.
Design details are given in \hyperref[sec:methods]{Methods}.

\subsection{Experiments}
We evaluate HEP on three materials-science research tasks, each posing
an open research question of the same form: ``which crystal structure
does a single fixed machine-learned interatomic potential (MLIP) prefer
for each compound in a family, and what physics or chemistry explains
the pattern?''
Task AO$_2$ concerns polymorph selection among seven prototypes in the
AO$_2$ dioxides; task A$_2$BB$'$O$_6$ concerns B-site cation ordering
in the A$_2$BB$'$O$_6$ double perovskites, among rock-salt, layered,
and columnar orderings and a disordered reference; and task MX
concerns prototype selection among rocksalt, NiAs, zincblende, and
wurtzite in the MX transition-metal monochalcogenides and
monopnictides.

The three families span combinatorial complexity from binary (AO$_2$,
MX) to quaternary (A$_2$BB$'$O$_6$) compounds, and the questions they
pose demand explanation rather than optimization toward a predefined
target.
The LLM has some prior knowledge of these systems, but the behavior
of the fixed potential is not known until it is actually tested.
Because the aim is new knowledge rather than the recall of known
facts, each hypothesis the agent forms can be judged only by testing
it.
Individual preferences can be computed by relaxation, but the rule
that governs the pattern cannot be read off by enumeration alone.
The questions are also scientifically meaningful, in that
understanding which crystal structure a potential stabilizes, and why,
bears directly on materials design.

In each run, the agent receives a single task prompt and is required
to work out an answer to the research question autonomously with the
provided tools and to deliver it as a final report.
Unless otherwise noted, the base LLM is GPT-5.5~\cite{OpenAI2026GPT55}.
The agent setup and
run settings are given in \hyperref[sec:methods]{Methods}.

\begin{figure}[!htb]
    \centering
    \includegraphics[width=\textwidth]{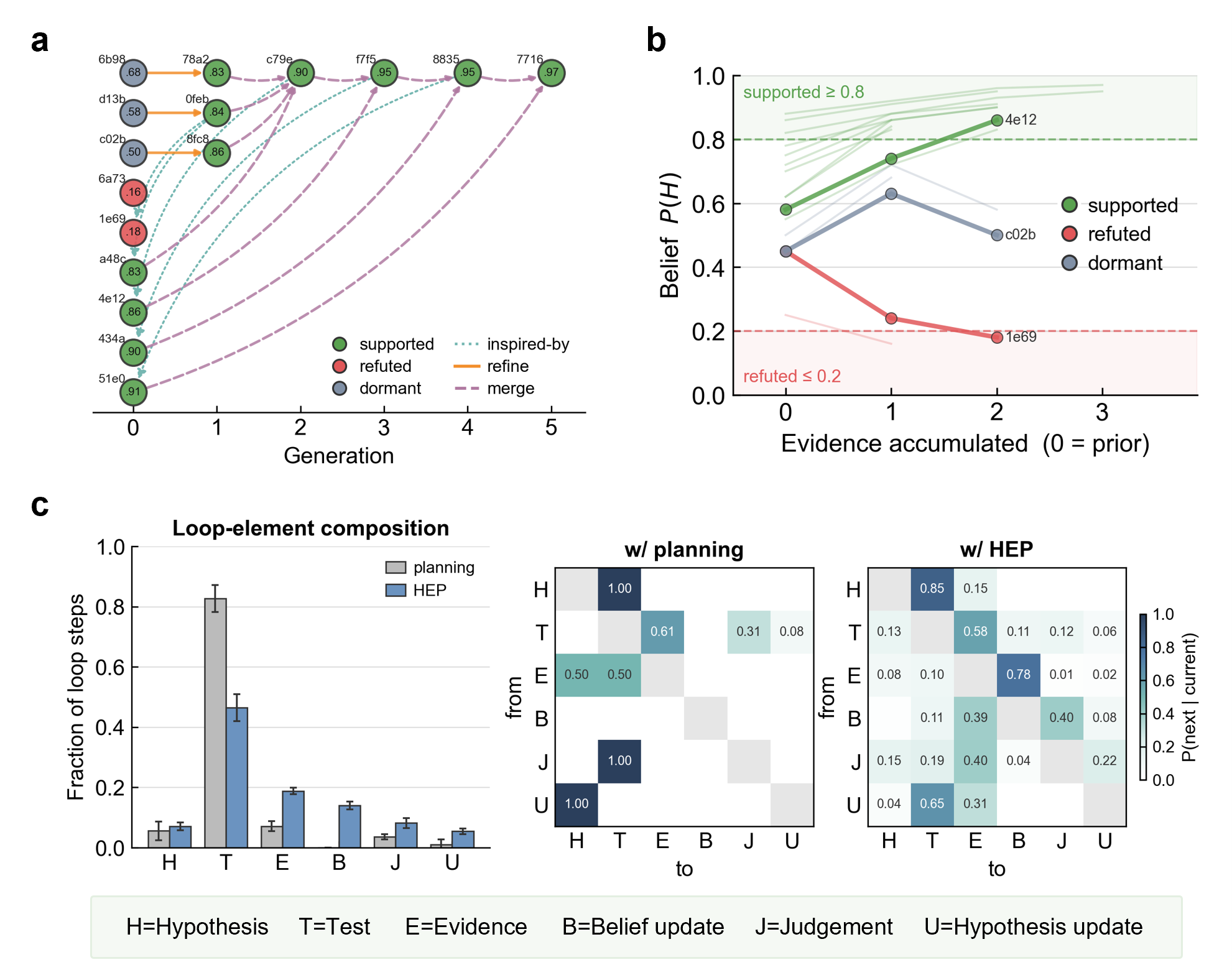}
    \caption{%
        \textbf{HEP makes the scientific loop explicit and auditable.}
        \textbf{(a)} Hypothesis tree of an example run on task AO$_2$.
        Nodes are hypotheses (fill = final lifecycle state, number = final
        belief, label = id suffix); edges show the generation mechanism
        (orange solid = \emph{refine}, purple dashed = \emph{merge}, teal
        dotted = \emph{inspired-by}, which contributes provenance but not
        lineage).
        Generation is 0 for \emph{de-novo} and \emph{inspired-by}
        hypotheses, parent$+1$ for \emph{refine}, and
        $\max(\text{parents})+1$ for \emph{merge}.
        \textbf{(b)} Belief trajectories for all 16 hypotheses (light lines)
        against the number of accumulated validated-evidence updates
        (0 = prior at proposal).
        One example per final state is highlighted.
        Shaded bands mark the registry-enforced verdict thresholds
        (supported $\ge 0.8$, refuted $\le 0.2$).
        \textbf{(c)} The scientific loop of a planning-style agent
        (baseline) versus the HEP-equipped agent
        (same goal, same base LLM, $n=3$ runs each).
        Left: fraction of loop steps by element (H = hypothesis proposal,
        T = test,
        E = evidence, B = belief update, J = judgement, U = hypothesis
        update; mean $\pm$ standard deviation across runs).
        Middle and right: row-normalized transition probabilities
        $P(\text{next} \mid \text{current})$ averaged over runs, after
        compressing consecutive repeats of the same element.
        Self-transitions are therefore undefined and shown in gray.
        An all-white row means the element never occurred; rows H, J, and U
        of the planning-style agent contain only one to three events each, so the
        corresponding probabilities rest on few transitions.
    }
    \label{fig:hep_loop}
\end{figure}

\paragraph{HEP operates the hypothesis--test--evidence--belief cycle.}
We first look at how the agent operates HEP within a single research
run, taking a run on task AO$_2$ as an example.
Figure~\ref{fig:hep_loop}a shows the resulting hypothesis tree, in
which each node is a hypothesis, colored by its final lifecycle state
and labeled with its final belief, and edges indicate the generation
mechanism.
The agent enumerated sixteen candidate hypotheses, exercising all four
generation mechanisms within this single run, and every hypothesis was
eventually transitioned to \texttt{supported}, \texttt{refuted}, or
\texttt{dormant}, with none left open at the end.
Toward the end of the run, the candidates converged through merging.
A four-parent merge produced a candidate governing rule, which was
consolidated through further evidence-backed merges into a final
hypothesis holding a belief of $0.97$.
The tree thus makes it possible to trace how each hypothesis relates
to its parents and through which mechanism it was generated.

Figure~\ref{fig:hep_loop}b traces belief through the same run.
The vertical axis is the belief $P(H)$ of each hypothesis, and the
horizontal axis is the number of accumulated validated-evidence updates,
with 0 corresponding to the prior at proposal.
Light lines show all sixteen hypotheses proposed in the run, one
trajectory per final state is highlighted as an example, and the
shaded bands mark the verdict thresholds enforced by the registry.
Every belief change corresponds to a recorded evidence attachment that
passed the validation gate, so each trajectory can be read as a
complete audit trail from prior to final belief.
The highlighted examples illustrate the three kinds of outcome:
hypothesis \texttt{4e12}, whose evidence raised its belief across the
support threshold; hypothesis \texttt{1e69}, whose evidence drove its
belief below the refutation threshold; and hypothesis \texttt{c02b}, which
remained mid-range and was shelved as dormant after being superseded
by its refined child.
Verdicts occur only after belief crosses these thresholds, so every
verdict is grounded in validated evidence.

To examine how harnessing the LLM with HEP changes the behavior of the
agent, we compare against a planning-style agent
(plan--execute--replan~\cite{Yao2023ReAct,Sun2023AdaPlanner,Shinn2023Reflexion})
given the same research goal, the same tools for computation, and the
same base LLM, with $n=3$ runs per condition.
Following a recent analysis of agent execution
traces~\cite{RiosGarcia2026AIScientists}, we classify every step of
every run into hypothesis proposal (H), test (T), evidence (E), belief update
(B), judgement (J), or hypothesis update (U).
The left panel of Fig.~\ref{fig:hep_loop}c shows the fraction of steps
spent on each element, averaged over runs, and the middle and right
panels show the transition probabilities
$P(\text{next} \mid \text{current})$ between consecutive elements for
the planning-style agent and the HEP-equipped agent, respectively.
Because the two conditions stop under different criteria, we compare
these run-length-independent quantities rather than absolute counts.
The planning-style agent spends $83\%$ of its loop steps on tests and
$0\%$ on belief updates.
It moves from hypotheses to tests to evidence
(H$\rightarrow$T $1.00$, T$\rightarrow$E $0.61$) and then returns to
further hypotheses or tests, never explicitly externalizing a belief
and updating it against the collected evidence.
In our runs, at least, explicit belief revision did not emerge
spontaneously from planning alone, and eliciting it would require
dedicated support such as belief-oriented prompting.
Under HEP, in contrast, beliefs are explicitly externalized in the HEP
Registry, and the transition matrix exhibits the flow along the cycle,
H$\rightarrow$T ($0.85$), T$\rightarrow$E ($0.58$), E$\rightarrow$B
($0.78$), and B$\rightarrow$J ($0.40$), with judgements further
triggering hypothesis updates that are sent back to test
(J$\rightarrow$U $0.22$, U$\rightarrow$T $0.65$).
These transitions demonstrate that HEP drives the agent through the
full hypothesis--test--evidence--belief cycle.
The planning-style agent can also iterate a hypothesis--test cycle, but
it is difficult to trace on which evidence a belief was updated, how
that update led to a judgement, and on what basis each hypothesis was
generated.
HEP removes this ambiguity.
Every belief change is anchored to a specific piece of evidence, and
every hypothesis carries an unbroken record from proposal to verdict.

\FloatBarrier

\begin{figure}[!htb]
    \centering
    \includegraphics[width=\textwidth]{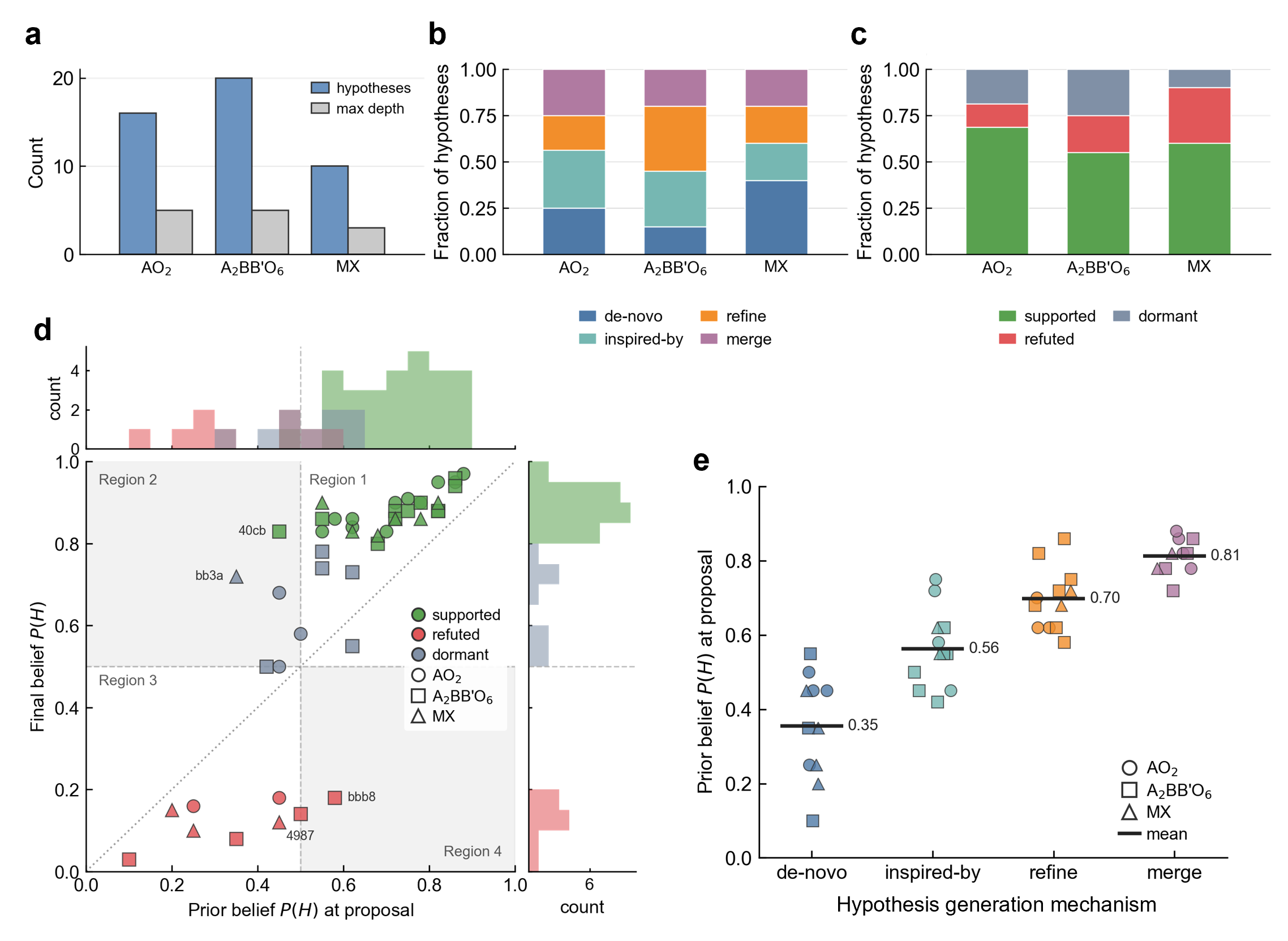}
    \caption{%
        \textbf{Behavior of the HEP-equipped agent across different
        research questions.}
        \textbf{(a)} Number of hypotheses generated and maximum generation
        depth per task.
        \textbf{(b)} Composition of the generation mechanisms of the
        hypotheses.
        \textbf{(c)} Final lifecycle states of the generated hypotheses.
        \textbf{(d)} Self-assigned prior belief $P(H)$ at proposal versus
        final belief for all hypotheses (color = final state,
        marker shape = task).
        Dashed lines at $0.5$ divide the plane into confirmation
        (Regions 1, 3) and belief-flip (Regions 2, 4) quadrants.
        Marginal histograms show the distributions of the prior belief
        $P(H)$ (top) and the final belief (right), colored by final
        state.
        \textbf{(e)} Prior belief at proposal grouped by generation
        mechanism (bars = group means).
    }
    \label{fig:hep_tasks}
\end{figure}

\paragraph{HEP generalizes across research questions.}
We next examine the generality of HEP across research questions,
applying it to task A$_2$BB$'$O$_6$ and task MX as well.
Figure~\ref{fig:hep_tasks} shows the overall results.
Across the three tasks the agent generated 10--20 hypotheses per task
with maximum generation depths of 3--5, and all four generation
mechanisms, \emph{de-novo}, \emph{inspired-by}, \emph{refine}, and
\emph{merge}, were exercised in every task
(Fig.~\ref{fig:hep_tasks}a,b).
Every hypothesis in all three tasks reached a terminal disposition, that
is, no hypothesis was left proposed or under test at the end of any run,
with \texttt{supported} being the most common final state in every task
(Fig.~\ref{fig:hep_tasks}c).

We then examine how the belief of each hypothesis changed from proposal
to verdict across the tasks.
Figure~\ref{fig:hep_tasks}d plots the self-assigned prior belief at
proposal against the final belief after all evidence for all
hypotheses, with the color and the marker shape indicating the final
state and the task, respectively.
The dashed lines at $P(H)=0.5$ divide the plane into four regions:
hypotheses in Regions 1 and 3 ended on the same side of $0.5$ as they
were proposed, whereas those in Regions 2 and 4 crossed to the opposite
side, that is, their beliefs were flipped by the evidence.
In the marginal histograms, the priors of eventually supported,
refuted, and dormant hypotheses largely overlap, whereas the final
beliefs separate cleanly by final state.
The verdicts were thus determined by the agent's interpretation of the
accumulated evidence rather than by its initial guesses.
Belief flips occurred in both directions and in all tasks
(Regions 2 and 4).
We note that no hypothesis proposed with a prior above $0.7$ was
eventually flipped.
Indeed, every such hypothesis ended supported.
This can be read in two ways: the agent assigned high priors only to
hypotheses that deserved them, or it did not test its high-prior
hypotheses severely enough to overturn them.

Finally, we characterize the agent's use of the four hypothesis
generation mechanisms by the prior beliefs it assigns at proposal.
In Fig.~\ref{fig:hep_tasks}e, each point is a hypothesis, grouped by
its generation mechanism, with the vertical axis showing the prior
belief at proposal; the marker shape indicates the task, and the
horizontal bars are the group means.
The mean prior rises monotonically from \emph{de-novo} ($0.35$) through
\emph{inspired-by} ($0.56$) and \emph{refine} ($0.70$) to \emph{merge}
($0.81$), and the distributions of \emph{de-novo} and \emph{merge} do
not overlap, with hypotheses from all three tasks mixed within each
group.
Although self-reported, the priors are internally consistent:
hypotheses built on more validated knowledge, such as \emph{refine} and
\emph{merge} children, receive systematically higher priors than
\emph{de-novo} proposals.

In each task the run culminated in a supported governing rule with high
final belief, verified by predicting the preferred structures of
compositions the agent had not yet examined when it formed the rule.
Table~\ref{tab:final_hypotheses} lists the final hypothesis of each
task, with its final belief taken from the registry and a one-line
summary condensed from its statement.
The histories behind these endpoints illustrate how belief flips shaped
the discoveries.
Hypothesis \texttt{40cb} on task A$_2$BB$'$O$_6$ (annotated in
Fig.~\ref{fig:hep_tasks}d), claiming that a secondary B-site
size/covalency contrast can still drive rock-salt order even without
charge contrast, was proposed skeptically at $0.45$, was driven to
$0.83$ by two validated tests, and became a parent of the first merge
of the task, ending as a pillar of the final rule.
Conversely, a textbook-plausible electronic mechanism on the same
task, Jahn--Teller/spin-state compatibility (\texttt{4987} in
Fig.~\ref{fig:hep_tasks}d), was proposed at $0.50$ and refuted to
$0.14$.
Its data-driven refinement, redox/covalency complementarity of
the B-site pair (\texttt{bbb8}), was proposed at $0.58$ and refuted in
turn to $0.18$.
These recorded refutations redirected the search toward the A-site
electronic exception that survives in the final rule.
All of these histories are read directly from the event logs in the
registry, an account that would be difficult to reconstruct from
unstructured agent logs.
We note, however, that the governing rules discovered here reflect the
behavior of the fixed MLIP selected by the agent, that the evidence
behind them consists of MLIP-level energetics and structural analyses
rather than electronic-structure calculations, and that their
correspondence to experimentally established materials physics remains
to be examined.

\begin{table}[!htb]
    \centering
    \caption{%
        Final supported hypothesis of each task.
        Final beliefs are taken from the HEP Registry; the governing
        rule is a one-line summary condensed from the hypothesis
        statement.
    }
    \label{tab:final_hypotheses}
    \small
    \begin{tabular}{lcp{9.4cm}}
        \toprule
        Task & Final belief $P(H)$ &
        Governing rule (one-line summary) \\
        \midrule
        AO$_2$ & $0.97$ &
        Prioritized hierarchy: molecule-forming C/S, then lone-pair
        Se/Te, then $f$-block fluorite, then cation-radius regimes. \\
        A$_2$BB$'$O$_6$ & $0.96$ &
        Charge contrast as the primary rule, refined by size and
        covalency, with A-site electronic exceptions
        (Ba/Pb--Cr/Co, Pb off-centering). \\
        MX & $0.90$ &
        Three-tier chemical hierarchy with NiAs (B8) as default,
        tetrahedral windows for $d^{10}$ and covalent
        sulfide/selenide chemistries, and a late-5$d$ pnictide
        override. \\
        \bottomrule
    \end{tabular}
\end{table}

\FloatBarrier

\begin{figure}[!htb]
    \centering
    \includegraphics[width=\textwidth]{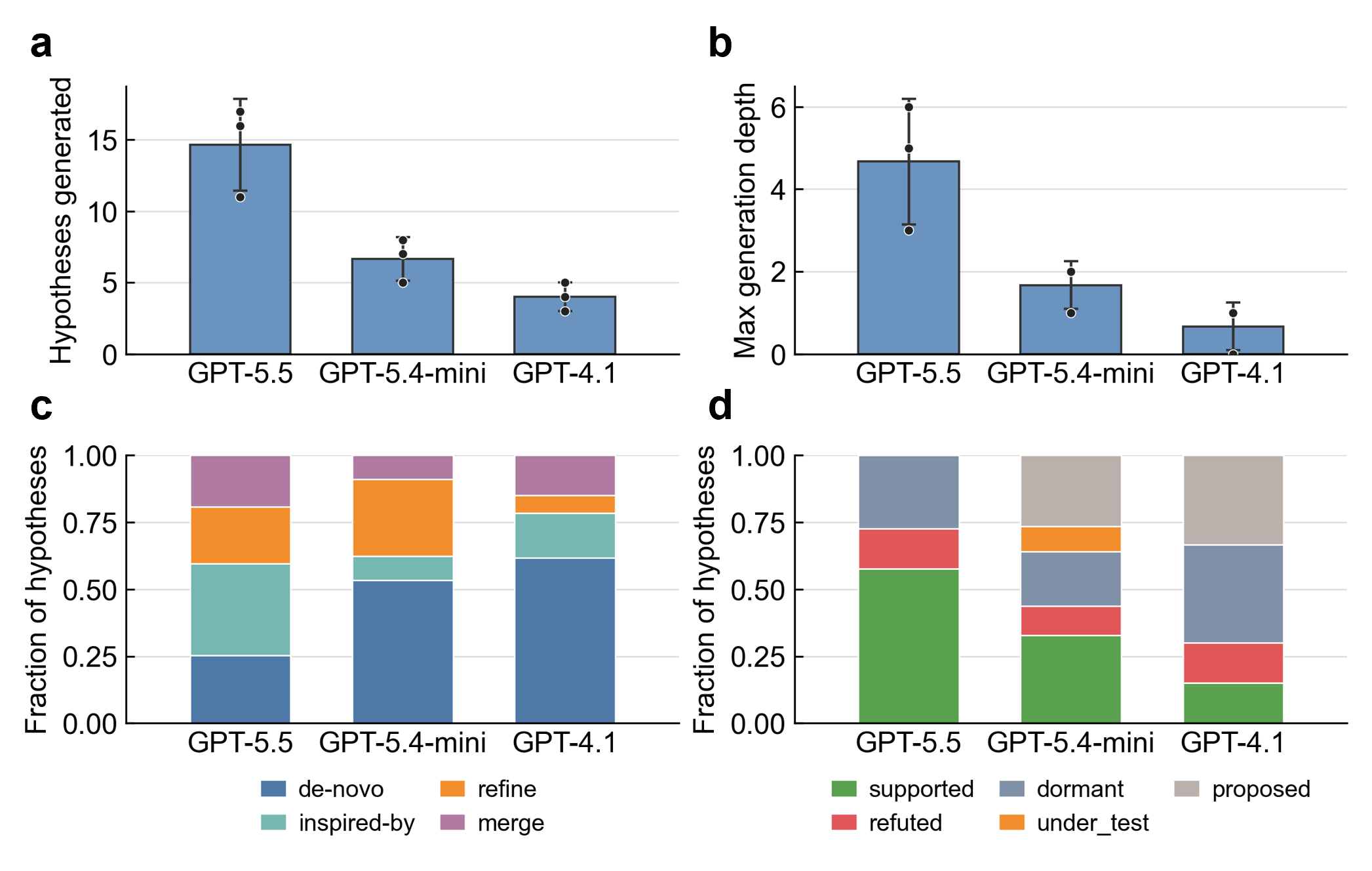}
    \caption{%
        \textbf{LLM dependence of the ability to operate HEP.}
        \textbf{(a)} Mean and standard deviation of the number of
        hypotheses generated per run on task AO$_2$ when only the
        base LLM is changed ($n=3$ runs per LLM; dots show individual
        runs).
        \textbf{(b)} Mean and standard deviation of the maximum generation
        depth.
        \textbf{(c)} Composition of the generation mechanisms of the
        hypotheses.
        \textbf{(d)} Final lifecycle states of the generated hypotheses;
        supported, refuted, and dormant
        are terminal (closed), while proposed and under-test indicate
        hypotheses left unresolved at run end.
    }
    \label{fig:hep_capability}
\end{figure}

\paragraph{HEP scales with base-LLM capability.}
HEP is an agent harness and can in principle be combined with any LLM
engine.
The ability to operate the protocol, however, should depend on the
reasoning and tool-use capabilities of the underlying LLM.
We therefore repeated task AO$_2$ with three base LLMs of
decreasing capability (GPT-5.5, GPT-5.4-mini~\cite{OpenAI2026GPT54Mini},
and GPT-4.1~\cite{OpenAI2025GPT41}) under
identical prompts, tools, and stopping conditions, with $n=3$ runs per
LLM (Fig.~\ref{fig:hep_capability}).
Figure~\ref{fig:hep_capability}a shows the mean and standard deviation,
over the three runs, of the number of hypotheses generated per run, and
Fig.~\ref{fig:hep_capability}b shows those of the maximum generation
depth reached in each run.
The mean count drops from $14.7$ to $6.7$ to $4.0$, and the mean depth
falls from $4.7$ to $1.7$ to $0.7$.
These results indicate that as LLM capability decreases, the agent
generates fewer hypotheses and becomes unable to grow generations of
hypotheses through refinement and merging.
GPT-4.1, for example, either leaves its hypotheses entirely unevolved
or grows them by only a single generation.

The composition of generation mechanisms shifts accordingly
(Fig.~\ref{fig:hep_capability}c).
GPT-5.5 uses the full repertoire in balanced proportions, whereas the
weaker LLMs increasingly default to \emph{de-novo} proposals, whose
share rises from 25\% for GPT-5.5 to 53\% and 62\% for the weaker two.
The same gradient appears in the final states of the hypotheses
(Fig.~\ref{fig:hep_capability}d).
GPT-5.5 resolved every hypothesis in all runs, whereas GPT-5.4-mini left
36\% of hypotheses open on average, and GPT-4.1 was unstable from run to
run.
Two of its runs resolved everything, while the third proposed three
hypotheses and abandoned them all, running computations but never
attaching their results to any hypothesis as evidence.

These results indicate that operating the individual HEP tools is
within the reach of every LLM tested, but that a more capable LLM
commands HEP as a whole, exercising the full set of tools and
continuing to evolve its hypotheses while carrying each proposed
hypothesis to a verdict.
The value of HEP is therefore realized most fully with capable base
LLMs.

\FloatBarrier

\section{Conclusions}

This work proposes the Hypothesis Evolution Protocol, an agent harness
that provides hypothesis generation, evaluation, and evolution as
explicit, auditable operations, making each hypothesis a persistent
object whose belief is moved on the basis of the evidence attached to
it.
On materials-science research tasks, the HEP-equipped agent operated
the full hypothesis--test--evidence--belief cycle that a planning-style
agent with the same base LLM left unexercised, and the protocol
generalized across three research questions, with each run culminating
in a supported governing rule verified by predicting the preferred
structures of compositions the agent had not yet examined.
How fully the protocol was exploited scaled with the capability of the
base LLM.
HEP does not substitute for that capability, in line with a
recent analysis arguing that the quality of scientific reasoning is
dominated by the base LLM rather than by the scaffold around
it~\cite{RiosGarcia2026AIScientists}.
Harness and capability are instead complementary.
The harness determines whether scientific reasoning is externalized
and verifiable, and capability determines how fully it is exploited.

This work examined the hypothesis--test--evidence--belief cycle in a
single-agent setting, in which the agent itself attaches evidence and
assesses its validity.
A natural next step is to make the validation of evidence more
rigorous, for example by replacing the agent's self-assessment with
programmatic checks where the tests permit, or by introducing an
independent auditor agent that reviews attached evidence and belief
updates.
Because HEP externalizes every hypothesis, evidence, and belief update
as a persistent record, such external validators can operate directly
on the existing registry, moving the cycle from self-reported toward
independently verified.
Likewise, the present study operated at the fidelity of a fixed MLIP;
equipping the agent with first-principles tools such as density
functional theory would allow the evidence behind each belief update
to be validated more accurately.
As base LLMs continue to advance, auditable harnesses such as HEP will
turn that growing capability into scientific reasoning that human
researchers can inspect, verify, and build upon, bringing AI agents a
step closer to being trustworthy partners in scientific discovery.

\section{Methods}\label{sec:methods}

\subsection{HEP implementation}
The HEP Registry is an append-only, event-sourced store.
Every operation appends an event that is stamped with its author and
hash-chained to the previous one, no event is overwritten or deleted,
and the current lifecycle state and belief of a hypothesis are derived
by replaying its event log, which is persisted in the run's session
directory.
Beyond the fields introduced in Results, each hypothesis records a
testable predicted observable and its lineage (parent hypotheses and
generation).
An \emph{inspired-by} link is recorded as provenance only
and does not make the new hypothesis a descendant.

The agent operates the registry through seven tools.
\texttt{propose\_hypothesis} registers a new hypothesis under a
research question, and \texttt{transition\_hypothesis} moves a
hypothesis between lifecycle states.
\texttt{attach\_evidence} records one piece of evidence on a
hypothesis and updates its belief from it.
\texttt{list\_hypotheses} returns the competing set grouped by
research question, with the current state and belief of each
hypothesis, and \texttt{get\_hypothesis} returns the full record of a
single hypothesis, including its event timeline.
Evolution is evidence-driven by construction:
\texttt{refine\_hypothesis} creates a child hypothesis while the
parent retains its own state, \texttt{merge\_hypotheses} combines
hypotheses on the same question and transitions unresolved parents to
\texttt{dormant} as subsumed, and both operations are accepted only
for parents that have reached a verdict or have at least one piece of
evidence attached, so that untested proposals cannot produce
descendants.

Beliefs are elicited rather than computed.
The agent states a prior
$P(H)$ with a written rationale at proposal, and at each evidence
attachment (with kind simulation, experiment, literature, derivation,
or analysis, and direction supports, refutes, or inconclusive) it
states an updated $P(H)$, recorded with its rationale and the
identifier of the triggering evidence.
The two belief rules described in Results are enforced by the tools
rather than left to the agent.
For computational and analytical evidence, the belief moves only if
the agent certifies the result as trustworthy for its method, with a
diagnostic covering both the adequacy of the setup and the validity of
the result.
An uncertified attachment is recorded but marked
insufficient.
Verdict transitions whose belief has not crossed the threshold, and
any move outside the lifecycle state machine, are rejected.

\subsection{Agent setup}
All experiments use a single autonomous research agent, implemented
with the OpenAI Agents SDK, that receives one task prompt and runs to
completion without human input.
Besides the HEP tools, the agent has computational tools for
atomistic materials research exposed as MCP servers, including
MLIP-based structure relaxation, molecular dynamics, and
materials-database queries, together with a library of documented
skills it can discover and execute, both provided by
AtomisticSkills~\cite{Deng2026AtomisticSkills}, as well as a tool for
authoring and executing its own analysis code, with each script
retained in the run's record, and recording tools for logging findings
and writing the final report.
The agent runs in an outer loop.
Each iteration allows up to 80 LLM
turns, after which the driver re-prompts it to continue, appending the
remaining wall-clock budget so that the agent can select tests that
fit within the remaining time.
The conversation history is persisted,
and every run is given a 24-hour wall-clock budget, which no run
exhausted.
The HEP-equipped agent is given the HEP tools and a
hypothesis-discipline prompt but no planning tools.
The run ends when
the investigation saturates, that is, when no tool call is made for
three consecutive iterations, at which point the agent is prompted to
consolidate its findings and write the final report.

\subsection{Trajectory analysis}
For the loop analysis in Fig.~\ref{fig:hep_loop}c, each step of a run
is classified into the loop elements.
HEP tool calls map deterministically: \texttt{propose\_hypothesis} to
H; \texttt{refine\_hypothesis} and \texttt{merge\_hypotheses} to U;
\texttt{attach\_evidence} to E, and additionally to B when the
attachment moved the belief; and \texttt{transition\_hypothesis} to J
for verdict transitions.
Calls to computational tools map to T, free-text steps are classified
by an LLM (GPT-5.5), and scaffolding steps such as file management are
excluded.
Consecutive repeats of the same element are compressed, and the
transition matrices count adjacent pairs of elements, row-normalized
and averaged over runs.

\section*{Acknowledgments}
This work was supported by ACT-X (grant no. JPMJAX24DB) and BOOST (grant no. JPMJBS2418), Japan Science and Technology Agency, Japan. 


\section*{Code Availability}
The source code developed for this study will be made publicly
available upon publication.

\bibliographystyle{unsrt}
\bibliography{references}

\end{document}